\documentclass[11pt,a4paper]{article}
\usepackage{times,latexsym}
\usepackage{url}
\usepackage[T1]{fontenc}
\usepackage{graphicx}
\usepackage{todonotes}
\usepackage{booktabs}
\usepackage{multirow}
\usepackage[colorlinks=true,pagebackref=true]{hyperref}
\usepackage[capitalise,noabbrev,nameinlink]{cleveref}
\usepackage{amssymb}% http://ctan.org/pkg/amssymb
\usepackage{pifont}% http://ctan.org/pkg/pifont
\usepackage{booktabs}
\usepackage[usestackEOL]{stackengine}
\newcommand{\cmark}{\ding{51}}%
\newcommand{\xmark}{\ding{55}}%
\usepackage[acceptedWithA]{tacl2021v1}
\usepackage{xspace,mfirstuc,tabulary}

\newif\iftaclinstructions
\taclinstructionsfalse % AUTHORS: do NOT set this to true
\iftaclinstructions
\renewcommand{\confidential}{}
\renewcommand{\anonsubtext}{(No author info supplied here, for consistency with
TACL-submission anonymization requirements)}
\newcommand{\instr}
\fi

\iftaclpubformat % this "if" is set by the choice of options

\else

\fi

\title{Benchmarking Large Language Models for News Summarization}

\author{
  Tianyi Zhang$\phantom{}^{1}$\Thanks{Equal Contribution. Order determined by a random coin flip. Correspondence to \url{tz58@stanford.edu} and \url{faisal@cs.columbia.edu}}, Faisal Ladhak$\phantom{}^{2*}$, Esin Durmus$\phantom{}^{1}$, Percy Liang$\phantom{}^{1}$,
  \\
  \textbf{Kathleen McKeown$\phantom{}^{2}$, Tatsunori B. Hashimoto$\phantom{}^{1}$}
  \\
  $\phantom{}^{1}$Stanford Univeristy
  $\phantom{}^{2}$Columbia Univeristy
  \\
}

\date{}

\begin{document}
\maketitle

% move to intro
% 1. Metrics are bad with old references
% 1.1. Still correlates decently with relevance
% 1.2. still correlate within setting/LLM family
% 2. Metrics are slighter better with our references but still make important mistakes. i.e. BRIO is really still at the top.
% 3. Reference-free metrics might be better

% Even with better reference summaries, automatic metrics are still not effective in evaluating summaries generated by LLMs, which calls for future work.

\begin{abstract}
%KMFINAL - I am going to directly reword this to save space and I think it will make the point better, but I am saving everything in comments so you can revert back. 
Large language models (LLMs) have shown promise for automatic summarization but the reasons behind their successes are poorly understood.
By conducting a human evaluation on ten LLMs across different pretraining methods, prompts, and model scales,  we make two important observations.
% have promising potential for automatic summarization but their success factors are poorly understood.
%KMFINAL - You use key below
%key 
First, we find instruction tuning, and not model size, is the key to the LLM's zero-shot summarization capability.  Second, existing studies have been limited by low-quality references, leading to underestimates of human performance and lower few-shot and finetuning performance.
%Even the largest LLM cannot summarize well in the zero-shot setting without instruction tuning.
%Second, current studies are hindered by the low-quality reference summaries in existing benchmarks
%and this causes
%KMFINAL cut
%Systems that take supervision either through 
%finetuning and few-shot prompting approaches to learn the undesired behaviors in the references. 
%KMFINAL
%summaries. 
% \kmnote{Do you need the second part of this sentence? It's not exactly clear and makes the abstract long. Couldn't you just stop at the comma? }
% which gives an advantage to zero-shot LLMs and makes comparison difficult.
To better evaluate LLMs,
%KMFINAL
%we collect high-quality summaries from freelance writers.
we perform human evaluation over high-quality summaries we collect from freelance writers.
Despite major stylistic differences such as the amount of paraphrasing, we find that LMM summaries are judged to be on par with human written summaries.
\end{abstract}

%LLMs have shown remarkable success in zero-shot and few-shot performance on a large number NLP tasks. 
% 

% In this work, we explore the effectiveness of LLMs for news summarization and compare zero-shot and few-shot performance against state-of-the-art finetuned models. 
% We find that automated metrics tend to overwhelmingly prefer the summaries generated by current SOTA finetuned models over the zero-shot/few-shot summaries generated by LLMs. 
% Human evaluations, however, tell a different story -- human raters tend to overwhelmingly prefer the summaries generated by LLMs over those generated by finetuned models. 
% We find that the references in current summarization datasets are the culprit for this discrepancy, as increased exposure to training references leads to worse performance on human evaluations. 
% We further collect high-quality reference summaries from professional writers to better understand the quality of zero-shot LLM summaries. 
% We find that while zero-shot summaries are better than current SOTA summarizers, they are still largely extractive and considered worse than actual human-written summaries XX\% of the time. 
% Even with improved reference summaries, current evaluation metrics still do not correlate well with human judgements and we need further work on improving summarization metrics. 

\section{Introduction}
% Automatic summarization has long been an important target of text generation research.
Large language models (LLMs) have shown promising results in zero-/few-shot tasks across a wide range of domains~\citep{palm, anthropic, gpt3, opt} and raised significant interest for their potential for automatic summarization~\citep{gpt3-era, Liu2022RevisitingTG}. 
However, the design decisions contributing to its success on summarization remain poorly understood, and while prior work has shown that LLMs outperform prior state of the art, it remains unclear whether their outputs are comparable to human writers.
% and it is also unclear whether state-of-the-art LLMs are comparable to human summary writers.
Examining these questions is crucial for advancing future research in automatic summarization.
% Both questions are important to study to guide future research directions on automatic summarization.

% Our work aims to evaluate the effectiveness of LLMs on news summarization benchmarks, CNN/DM~\citep{cnndm} and XSUM~\citep{xsum}.

\begin{figure}
    \centering
    \includegraphics[width=\linewidth]{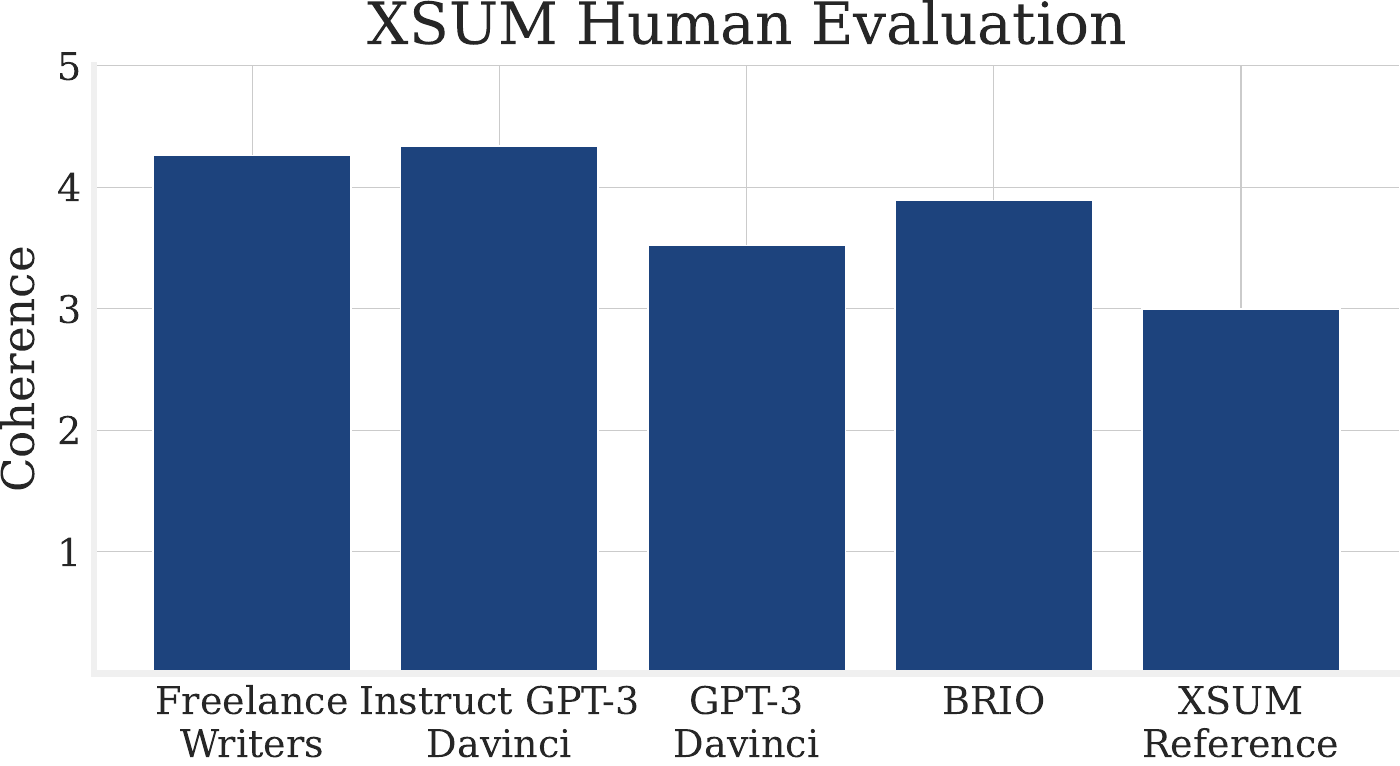}
    \caption{Selected annotator ratings of summary coherence on a 1 to 5 Likert scale.}
    \label{fig:fig1}
    \vspace{-15pt}
\end{figure}

% This work sets out to answer these two questions by benchmarking ten diverse LLMs with human evaluation on news summarization.
To answer the first question, we perform a systematic evaluation of ten diverse LLMs with human evaluation on news summarization and
our evaluation identifies instruction tuning to be the key to zero-shot summarization capability.
In contrast, self-supervised learning alone cannot induce strong summarization performance in the zero-shot setting (\Cref{fig:fig1}).
In fact, even a 350M parameter instruction-tuned GPT-3 can perform on par with the 175B parameter GPT-3.

To benchmark LLMs, we evaluated on the standard CNN/DM~\citep{cnndm} and XSUM datasets~\citep{xsum}, but found that low-quality reference summaries caused several issues.
% To the question of whether the state-of-the-art LLMs are comparable with human writers, we find it difficult to answer using existing benchmarks, CNN/DM~\citep{cnndm} and XSUM~\citep{xsum}).
The reference summaries in these benchmarks are of such poor quality that human annotators judge them to be worse than the outputs of most automatic systems (\Cref{fig:fig1}).
When computing automatic metrics using these references, their poor quality reduces the correlation between metric results and human judgement.
Not only does this make evaluation difficult, but it also degrades the performance of systems that take supervision either through finetuning or few-shot prompting and makes comparison difficult.
% learn the undesired behaviors from the reference summaries, which gives zero-shot LLMs an advantage and makes comparison difficult.

To address the quality issues of reference summaries and better understand how LLMs compare to human summary writers, we recruit freelance writers from Upwork\footnote{\url{https://www.upwork.com}} to re-annotate $100$ articles from the test set of CNN/DM and XSUM.
% To tease out the confounding effect of low quality reference summaries, we conduct a pilot study to collect high quality summaries from professional writers by reannotating $100$ articles from the test set of CNN/DM and XSUM.
% Then, we compare the best performing LLM, Instruct Davinci, to the freelance writer and find that the Instruct Davinci summaries are much more extractive
Comparing the best performing LLM, Instruct Davinci, to the freelance writers, we find that the Instruct Davinci summaries are much more extractive.
By manually annotating the summarization operations~\citep{Jing2000CutAP} used in these summaries, we find that Instruct Davinci paraphrases much less frequently although it is able to combine copied segments coherently.

Given their stylistic differences, we recruit annotators to compare the Instruct Davinci summaries to those written by freelance writers.
On aggregate, we find that Instruct Davinci is rated as comparable to the freelance writers.
However, analysis of individual annotators reveals that each annotator has a varying and stable preference for either Instruct Davinci or the freelance writers.

Together, our work makes the following key contributions. 
First, we identify instruction tuning, instead of model scale, as the key to LLMs' summarization capability.
Second, we show that reference summaries used in XSUM are judged by humans to be worse than the best LLM generated summaries. 
Third, to address the issue of low quality references, we collect better quality summaries from freelance writers and we show that the best LLM is rated as comparable to Upwork freelance writers.
In combination, these results call into question recent claims made about LLM summarization.  
In particular, summarization progress cannot be measured using reference-based metrics applied on XSUM. 
Furthermore,  the question of whether fine-tuned, few-shot or zero-shot models perform better remains an open question due to the poor quality of training data. 
% These problems are somewhat less severe in CNN/DailyMail, but still hold to a lesser degree. 
%but still hold for aspects of evaluation. 
To encourage furture work on improved evaluations, we release the high-quality summaries written by freelance writers and the evaluation data on 18 model settings and two datasets as resources\footnote{\url{https://github.com/Tiiiger/benchmark_llm_summarization}}.

% Submission-specific rules
\section{Background and Related Work}
\subsection{News Summarization}
News summarization is the task of producing a concise paragraph that captures the main points of a news article and has been a core problem within the field of automatic summarization~\citep{Radev2002IntroductionTT, Rush_2015, nallapati-etal-2016-abstractive,see-etal-2017-get,chen-bansal-2018-fast,dong2019unified}.
In this work, we benchmark LLMs on news summarization to understand their potential for automatic summarization and focus on two popular news summarization benchmarks, CNN/DM~\citep{cnndm} and XSUM~\citep{xsum}.

These two benchmarks contain large scale data in the order of houndreds of thousands summaries but are created via ``incidental supervison''.
CNN/DM includes articles from the CNN and DailyMail websites as the source article and adapt the bullet point highlights that come with the website articles as reference summaries.
XSUM includes articles from BBC news and adapts the bolded sentence(s) that appear in the first paragraph as reference summaries.
As a result, the reference summaries in these datasets are known to have quality issues~\citep{maynez-etal-2020-faithfulness,kang-hashimoto-2020-improved}, motivating us to addresses these defects to improve LLM evaluation.
% \kmnote{I would cut the rest of this}
% and discuss their implication in later sections.

To contextualize the performance of LLMs, we mainly compare to previous state-of-the-art approaches that leveraged supervised finetuning~\citep{liu-lapata-2019-text, bart, pegasus, brio}. 
%Compared to LLMs, these finetuned LMs have two distinctive features.
%First, finetuned LMs usually have only hundreds of millions of parameters.
%Second, finetuning requires using training examples to update the parameters of pretrained models.
% Besides summarization datasets and modeling methods, 
% \kmnote{Cut the phrase above. }
Summarization evaluation is another active area of research.
Many automatic metrics have been proposed~\citep{Lin2004ROUGEAP,bert-score,Sellam2020BLEURTLR,durmus-etal-2020-feqa,maynez-etal-2020-faithfulness,deutsch-roth-2021-understanding} but they do not always correlate with human evaluation of summarization systems~\citep{fabbri2020summeval,durmus-etal-2022-spurious}.
In this work, we evaluate the effectiveness of automatic metrics for evaluating LLMs and show that the usefulness of reference-based evaluation is closely linked to the quality of the references.

% \tianyi{Shall we make a table about parameters count and other difference in pretraining?}
\begin{table*}[htp]
    \centering
    \footnotesize
    \begin{tabular}{lcccc}
    \toprule
     Model & Model Creator & \# Parameters & Instruction Tuning & Reference  \\
    \midrule
     GPT-3 davinci v1  & \multirow{3}*{OpenAI} & 175B & \multirow{3}*{\xmark} & \multirow{3}*{\citet{gpt3}}\\
     GPT-3 curie v1  & & 6.7B & \\
     GPT-3 ada v1  & & 350M & \\
     \midrule
     InstructGPT davinci v2 & \multirow{3}*{OpenAI} & 175B & \multirow{3}*{\cmark} & \multirow{3}*{\citet{instruct-gpt}}\\
     InstructGPT curie v1  & & 6.7B &  \\
     InstructGPT ada v1  & & 350M &  \\
     \midrule
     OPT 175B & Meta & 175B & \xmark & \citet{opt}\\
     \midrule
     GLM & \Centerstack{Tsinghua\\University} & 130B & \xmark & \citet{glm}\\
     \midrule
     Cohere xlarge v20220609 & Cohere & 52.4B & \xmark & \citet{cohere} \\
     \midrule
     Anthropic-LM v4-s3 & Anthropic & 52B & \cmark & \citet{anthropic} \\
    \bottomrule
    \end{tabular}
    \caption{List of large language models we benchmarked with human evaluation.}
    \vspace{-5pt}
    \label{tab:model_details}
\end{table*}

\subsection{Large Language Models}
LLMs~\citep{Bommasani2021OnTO, palm, gpt3} have two distinctive features over previous pretrained models.
% \kmnote{drop "as indicated by their name"}
First, LLMs have much larger scale in terms of model parameters and training data.
% \kmnote{I would not make a new paragraph here. You only have 2 sentences above}
Second, unlike previous pretrained models that require finetuning, LLMs can be prompted zero-shot or few-shot to solve a task.
In the zero-shot setting, prompting presents the LLMs with inputs (e.g. news articles) and a natural language instruction (e.g., ``summarize this news article in three sentences'') and solicit outputs by having LLMs generate answers directly. 
When few-shot training examples are available, LLMs have the ability to learn "in context". 
Incontext learning prepends training input-output pairs along with the same style of instruction to the testing input. 
% After conditioning on training examples, LLMs can achieve even stronger performance.
% without updating their parameters, i.e., by conditioning on training examples that are prepended in the language modeling context.

%Since the initial advent of LLMs, the most notable technique to improve the prompting performance is instruction tuning~\citep{Sanh2021MultitaskPT, Wang2022BenchmarkingGV, instruct-gpt}.
Recently, instruction-tuning has emerged as an effective way to improve LLM prompting performance~\citep{Sanh2021MultitaskPT, Wang2022BenchmarkingGV, instruct-gpt}.
%Instruction tuning reformulates a diverse set of natural language processing tasks into the prompting format and updates the parameters of LLMs through either supervised finetuning or reinforcement learning. 
In this approach, a diverse set of natural language processing tasks are reformulated into the prompting format and the LLM's parameters are updated for these tasks either through supervised finetuning or reinforcement learning.

Recent work~\citep{goyal-durrett-2020-evaluating} shows that the instruct-tuned GPT-3 Davinci model is better than finetuned LMs, but do not show the design decision that contribute to the improved performance.
% \kmnote{Above you have "incontext". here "in-context". Which is it? }
%\kmnote{I feel that this is a weak comparison (the however). Could you instead end the last sentence by noting what they don't do. like "LMs, but do not show whether this results from instruction tuning or model size". Then you could cut below to after the ","}
In our work, we carry out a more comprehensive benchmark on ten different LLMs, to understand the effect of model scale, incontext learning and instruction tuning.
Given that automatic metrics may not be reliable, we focus on human evaluation as our benchmarking method.
\section{Human Evaluation on News Summarization Benchmarks}
\label{sec:helm_benchmark}
\begin{table*}[htp]
\resizebox{\textwidth}{!}{

\begin{tabular}{llllllll}
\toprule
 &  & \multicolumn{3}{c}{CNN/Daily Mail} & \multicolumn{3}{c}{XSUM} \\
Setting & Models & Faithfulness & Coherence & Relevance & Faithfulness & Coherence & Relevance \\
\midrule
\multirow[c]{6}{*}{Zero-shot language models} & GPT-3 (350M) & $0.29$ & $1.92$ & $1.84$ & $0.26$ & $2.03$ & $1.90$ \\
 & GPT-3 (6.7B) & $0.29$ & $1.77$ & $1.93$ & $0.77$ & $3.16$ & $3.39$ \\
 & GPT-3 (175B) & $0.76$ & $2.65$ & $3.50$ & $0.80$ & $2.78$ & $3.52$ \\
 & Ada Instruct v1 (350M*) & $0.88$ & $4.02$ & $4.26$ & $0.81$ & $3.90$ & $3.87$ \\
 & Curie Instruct v1 (6.7B*) & $0.97$ & $\mathbf{4.24}$ & $\mathbf{4.59}$ & $\mathbf{0.96}$ & $4.27$ & $\mathbf{4.34}$ \\
 & Davinci Instruct v2 (175B*) & $\mathbf{0.99}$ & $4.15$ & $\mathbf{4.60}$ & $\mathbf{0.97}$ & $4.41$ & $\mathbf{4.28}$ \\
\midrule
\multirow[c]{10}{*}{Five-shot language models} & Anthropic-LM (52B) & $0.94$ & $3.88$ & $4.33$ & $0.70$ & $\mathbf{4.77}$ & $4.14$ \\
 & Cohere XL (52.4B) & $\mathbf{0.99}$ & $3.42$ & $4.48$ & $0.63$ & $\mathbf{4.79}$ & $4.00$ \\
 & GLM (130B) & $0.94$ & $3.69$ & $4.24$ & $0.74$ & $4.72$ & $4.12$ \\
 & OPT (175B) & $0.96$ & $3.64$ & $4.33$ & $0.67$ & $\mathbf{4.80}$ & $4.01$ \\
 & GPT-3 (350M) & $0.86$ & $3.73$ & $3.85$ & - & - & - \\
 & GPT-3 (6.7B) & $0.97$ & $3.87$ & $4.17$ & $0.75$ & $4.19$ & $3.36$ \\
 & GPT-3 (175B) & $\mathbf{0.99}$ & $3.95$ & $4.34$ & $0.69$ & $4.69$ & $4.03$ \\
 & Ada Instruct v1 (350M*) & $0.84$ & $3.84$ & $4.07$ & $0.63$ & $3.54$ & $3.07$ \\
 & Curie Instruct v1 (6.7B*) & $0.96$ & $\mathbf{4.30}$ & $4.43$ & $0.85$ & $4.28$ & $3.80$ \\
 & Davinci Instruct v2 (175B*) & $\mathbf{0.98}$ & $4.13$ & $4.49$ & $0.77$ & $\mathbf{4.83}$ & $\mathbf{4.33}$ \\
\midrule
\multirow[c]{2}{*}{Fine-tuned language models} & Brio & $0.94$ & $3.94$ & $4.40$ & $0.58$ & $4.68$ & $3.89$ \\
 & Pegasus & $0.97$ & $3.93$ & $4.38$ & $0.57$ & $4.73$ & $3.85$ \\
\midrule
Existing references & - & $0.84$ & $3.20$ & $3.94$ & $0.37$ & $4.13$ & $3.00$ \\
\bottomrule
\end{tabular}

}
\caption{
Human evaluation results for zero-shot and five-shot LLMs, finetuned LMs, and reference summaries. 
We bold the entries that are not statistically significantly different from the best numbers in each column.
% We bold best performing numbers that are statistically indistinguishable from each other under a paired bootstrap test.
}
\vspace{-5pt}
\label{tab:summarization-human-evaluation}
\end{table*}
In this section, we use human evaluation to systematically benchmark a diverse set of ten LLMs on news summarization. 
% We ask human annotators to score the 
% \kmnote{So you decided to drop importance?}
% \kmnote{To make the construction parallel, drop "the" before "coherence"}
% faithfulness, relevance, and coherence of the summaries generated by different models. 
% \kmnote{I find it somewhat surprising ot see results of the evaluation first. Also, you have said it before. So if you wanted to save space you could cut the next 2 sentences. I get why you did it as it does make the point again, but I think it's been made earlier and will be made later so OK to cut. }
We observe that instruction tuning is the key to strong summarization capability and low-quality reference summaries in current benchmarks may underestimate few-shot or finetuning performance.
% In addition, we find it difficult to contextualize LLMs with the performance of human writers because reference summaries in existing benchmarks have low quality.
\subsection{Experimental Setup}
\label{sec:helm_experimental_setup}
\paragraph{Data}
We conduct our human evaluation on CNN/DM and XSUM by sampling a hundred examples from each validation set respectively. 
For the few-shot incontext learning settings, we sample five examples from the training set to be the demonstration examples. 
Due to the limited context window, we sample five articles that are between 50 and 150 tokens in length according to the GPT-2 tokenizer.
% \kmnote{"not nonsensical"??? Is that what you meant to say? }
For XSUM, we find that a uniform sampling occasionally result in articles that are unreadable due to data preprocessing so we manually pick from the training set.

\paragraph{Model Details}
We consider ten LLMs across different pretraining strategies and model scales\footnote{We note that the training details of instruction-tuned GPT-3 models may differ from those mentioned in the publication and are inferred by us based on the API naming scheme.}.
\Cref{tab:model_details} lists the details of the LLMs we consider. 
% \kmnote{I think it would be good to have details on these differences. It will be helpful to others working on this. You could put in appendix and point to that. }
% \kmnote{This sentence is a little confusing in part because we first have to look at the table to understand. Still you say "benchmark all modesl in the five-shot setting but ..." does this mean you benchmark the six (I would specify six in the text to be clear. you could even repeat more of the table here as people may not look at the table right away. For example 3 instruct-GPT models and 3 GPT models) OpenAI modesl in both the five-shot setting AND the zero-shot setting and all the other models you don't benchmark in the zero-shot setting? or do you only benchmark OpenAI in the zero-shot setting? Needs some clarification}
Due to limited computational resources and model access, we benchmark all models in the five-shot setting but only benchmark three OpenAI GPT-3 models and three OpenAI instruction-tuned GPT-3 models in the zero-shot setting.

For CNN/DM, we solicit LLM summaries with the following prompt template ``\texttt{Article: [article]. Summarize the article in three sentences. Summary:}''
For XSUM, we modify the prompt template to \texttt{summarize in one sentence} to match the style of the reference summaries.
For all LLMs we consider, we sample with temperature $0.3$ following prior work \cite{wu2021recursively}.

To contextualize our LLM benchmarking results, we also evaluate two state-of-the-art finetuned LMs: Pegasus~\citep{pegasus} and BRIO~\citep{brio}. 
% We evaluate the model checkpoints\footnote{for example, see \url{https://huggingface.co/google/pegasus-cnn_dailymail}} that are released on the huggingface model hub.
We decode the finetuned LMs using a beam size of $5$ following prior work \cite{bart}.
In addition, we also evaluate the existing reference summaries in the CNN/DM and XSUM validation sets. 
% \underline{GPT-3 \{ada, curie, davinci\}} are OpenAI models that respectively have 350M, 6B, and 175B parameters. In addition, we evaluate \underline{InstructGPT \{ada v1, curie v1, davinci v2\}}. These models are trained with instruction tuning although the exact training details may differ from the publication. We assume these models to have the same parameter count with the models that are not instruction tuned based on the naming scheme.

% We consider two open-sourced LLMs, GLM 130B and OPT 175B. 

\paragraph{Human Evaluation Protocol}
We recruit annotators from Amazon Mechanical Turk, compensating them at California minimum wage of \$15.00/hr using conservative time estimates as recommended by \citet{Whiting2019FairWC}. 
% \tianyi{@faisal: one sentence about the white-list pool of annotators.}
Each model summary was evaluated by three annotators and we report results based on their average score for each summary.

Our annotators evaluate each summary based on three criteria: faithfulness, coherence, and relevance.
We define these terms and collect data according to the guidelines in \citet{fabbri2020summeval}.
% Detailed definitions can be found in the same reference.
% We ask the annotators to evaluate each summary for three aspects:
% faithfulness, coherence, and relevance, following the guidelines of \citet{fabbri2020summeval} and 
% We refer to \citet{fabbri2020summeval} for detailed definitions of these aspects.
Coherence and relevance ratings are collected on a 1 to 5 Likert scale while faithfulness ratings are collected in binary ratings due to its binary nature.
% \kmnote{"Different from" -> "Unlike" (not fluent as is)}
Unlike \citet{fabbri2020summeval}, we omit evaluating fluency because we find LLM outputs to be mostly fluent.
The full annotation guidelines are included in our code release.
% We define a summary to be faithful if “all the information expressed by the summary can be inferred from the article” and solicit binary decisions from the annotators. We define a summary to be relevant if the summary “includes only important information from the source document” and coherent if the summary “organizes the relevant information into a well-structured summary”. For relevance and coherence, we ask the annotators to annotate on a 1-5 Likert scale.

% \kmnote{Probably you should say something about what you did to discourage spammers. Can you put your actual annotation guidelines in the appendix and point to them from the main paper? }

\subsection{Evaluation Results}
\label{sec:helm_results}
\begin{figure*}[ht]
    \centering
    \includegraphics[width=0.8\textwidth]{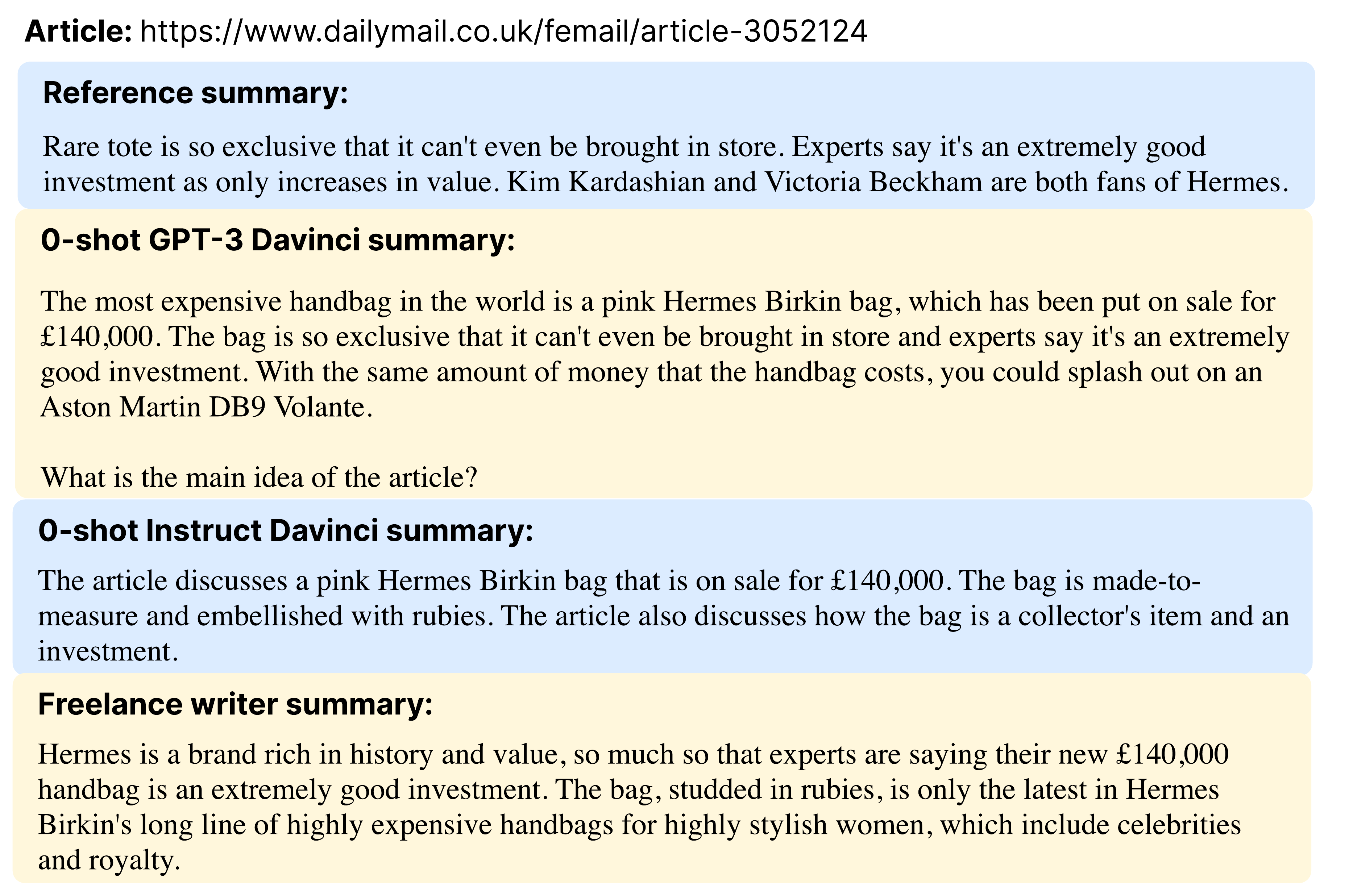}
    \caption{Examples summaries generated by GPT-3 models (\Cref{sec:helm_benchmark}) or written by freelance writers (\Cref{sec:upwork_written_summaries}) of an article from the CNN/DM dataset. We find that instruction-tuned GPT-3 model can generate a much better summary compared to the non-instruction-tuned variant. 
    Reference summary from CNN/DM is not coherent whereas
    freelance writer summary both coherent and relevant.}
    
    \label{fig:qualitative_example}
\end{figure*}

\Cref{tab:summarization-human-evaluation} presents the evaluation results\footnote{We note that the 350M GPT-3 consistently generates empty outputs so we omit it from the human evaluation.}.
We now discuss two main observations. 

\paragraph{Instruction tuned models have strong summarization ability.}
Across the two datasets and three aspects, we find that the zero-shot instruction-tuned GPT-3 models, especially Instruct Curie and Davinci, perform the best overall.
Compared to the fine-tuned LMs (e.g. Pegasus), Instruct Davinci achieves higher coherence and relevance scores (4.15 vs. 3.93 and 4.60 vs. 4.40) on CNN and higher faithfulness and relevance scores (0.97 vs. 0.57 and 4.28 vs. 3.85) on XSUM, which is consistent with recent work~\citep{gpt3-era}.
In contrast to instruction tuning, we find scale to be less important.
Even the largest 175B model often ignores the instruction and generates irrelevant content while the much smaller Instruct Ada outperforms the 175B GPT-3 model on coherence and relevance.

In the five-shot setting, non-instruction-tuned LLMs can improve their summarization performance through 
% \kmnote{Not consistent with how you spelled it above}
incontext learning.
% \kmnote{Either drop "score" or use plural. If you keep in the word "scores" I think it would read better at the end of the sentence: "as the instruction-tuned LLMs on faithfulness scores on CCN/DM and...". "similarly well" also not particularly fluent. I think it should just be "perform as well as.." or could be "perform similarly to.."}
For faithfulness scores on CNN/DM and coherence scores on XSUM, several non-instruction-tuned LLMs can perform as well as the instruction-tuned LLMs.
However, for other aspects, we still find the instruction-tuned LLMs to be better.

\paragraph{Reference summaries in current benchmarks are extremely low quality.}
We arrive at this conclusion based on two observations.
First, most automatic summarization systems score better than the reference summaries across all three aspects.
% \kmnote{Note spelling of in-context}
Second, applying incontext learning with the current reference summaries makes instruction-tuned models generate worse summaries.
For example, on the XSUM dataset, after conditioning on five reference summaries, the faithfulness score of Instruct Davinci drops from 0.97 to 0.77.
% \kmnote{"Besides" not right here. -> "In addition to", or perhaps "Aside from" ... though that not quite right either. Think about the right connective here. }

The low-quality reference summaries make it difficult to compare LLMs to both fine-tuned models and humans.
% \kmnote{I think you need to qualify this with "instruction-tuned" as otherwise it's confusing. You just said that zero-shot models don't perform well at all}
% Currently, we observe that zero-shot instruction-tuned LLMs perform best.
When comparing to finetuned models, the poor performance of fine-tuned models can be attributed to the low-quality references in training data and we may be underestimating the finetuning performance.
% However, these results suggest that the low performance of fine-tuned models may be partially due to the low-quality data used to train them.
When comparing to human, the low-quality references are not representative of human performance because they are created through heuristics. 
As a result, it's likely that the differences between instruction-tuned LLMs and human performance are likely overstated in \Cref{tab:metrics}.
\begin{table*}[htp]
\begin{tabular}{lcccccc}
\toprule
\small
& \multicolumn{3}{c}{CNN/DailyMail} & \multicolumn{3}{c}{XSUM} \\
Metric & Faithfulness & Coherence & Relevance & Faithfulness & Coherence & Relevance \\
\midrule
Rouge-L & 0.54 & 0.48 & 0.72 & -0.27 & 0.71 & 0.30 \\
METEOR & 0.58 & 0.37 & 0.66 & -0.22 & 0.68 & 0.38 \\
BertScore & 0.54 & 0.47 & 0.70 & -0.23 & 0.70 & 0.30 \\
BARTScore & 0.56 & 0.34 & 0.65 & -0.22 & 0.70 & 0.35 \\
BLEURT & 0.56 & 0.62 & 0.81 & -0.08 & 0.67 & 0.41 \\
% Lite Pyramid \\
% SummaC & 0.54 & 0.11 & 0.26 & 0.26 & -0.41 & -0.29 \\
% QAFactEval & 0.64 & 0.16 & 0.35 & 0.55 & 0.16 & 0.37\\
\bottomrule
\end{tabular}
\caption{
System-level kendall's tau correlation with human scores across different axes.
}
\label{tab:metrics}
\end{table*}

\paragraph{Qualitative Examples.}
\Cref{fig:qualitative_example} showcases example summaries on an article from the CNN/DM validation set, comparing the summaries of zero-shot GPT-3 Davinci, instruction-tuned GPT-3 Davinci, and the CNN/DM reference summary.
% \Cref{fig:qualitative_example} includes example summaries of an article from the CNN/DM validation set and we compare zero-shot GPT-3 Davinci, instruction-tuned GPT-3 Davinci, and the CNN/DM reference summary.

We start by noting that the zero-shot GPT-3 model cannot follow the instruction to summarize well. 
After the summary paragraph, the model generates an additional question that is completely irrelevant.
In addition to the failure of instruction following, the generated summary contains a factual error, stating that the handbag mentioned is the most expensive in the world, which contradicts the original article.
% the generated summary starts with a factual error: the said handbag is not the most expensive handbag in the world according to the original article.
In contrast, the instruction-tuned GPT-3 model generates a summary that is both faithful and coherent.

% \kmnote{Could you put a lot of examples in the appendix and point to them here? }
%KMFINAL -  cut a few words to get rid of the widow. 
We 
%can 
also observe from \Cref{fig:qualitative_example} that the reference summary is not coherent.
The brand ``Hermes'' is not introduced until the end and its connection to the rest of the story is unclear. 
% The brand ``Hermes'' is not mentioned until the end of the summary and it is unclear how this brand is related to the rest of the news story.
This is unsurprising as reference summaries in the CNN/DM dataset were originally bullet points accompanying the articles
%KMFINAL - check that you're OK with this rewroding. It saves a line. 
%not meant to form 
as opposed to a coherent paragraph.
% This may not be surprising as reference summaries in the CNN/DM dataset are original bullet points that appear along with the articles and are not meant to constitute a coherent paragraph.

\subsection{Understanding Automatic Metrics}
\label{sec:metric_1}
% \kmnote{This section feels very long and the story is not clear. Here is where you "waffle". You start Section 4 off with a very crisp statement but that does not come out in this section. I think if you shorten, it could. For example, could you cut the first two sentences below? Just start iwth "We compute"?}
% Recent work~\citep{gpt3-era} reports that current automatic metrics mistakenly prefer finetuned LMs to LLMs and do not capture human preferences.
% Using the human evaluation results in \Cref{sec:helm_results}, we expand on previous findings to understand if automatic metrics are helpful for comparing 
% LLMs. 
% \kmnote{either "comparing LLMs" or "comparisons among LLMs". Former is simpler and thus, probably better.}
% In this section we want to understand how well automated metrics are at evaluating LLM summarizers and whether they agree with human raters on summarization quality.
% Given the strong performance 
We compute six popular automatic metrics and compute their system-level correlations against human ratings.
The list of metrics we evaluate are: Rouge-L~\citep{Lin2004ROUGEAP}, METEOR~\citep{Banerjee2005METEORAA}, BertScore~\citep{bert-score}, BLEURT~\citep{Sellam2020BLEURTLR}, and BARTScore~\citep{Yuan2021BARTScoreEG}.
% Lite Pyramid~\citep{Shapira2019CrowdsourcingLP}
% , and two reference-free metrics, SummaC~\citep{Laban2021SummaCRN} and QAFactEval~\citep{fabbri-etal-2022-qafacteval}. 
% To do this, we compute system-level metric scores for each of the systems and compare this against human judgements.

\autoref{tab:metrics} shows Kendall's tau rank correlations between automated metrics and human judgements. 
We observe significantly different trends on CNN/DM and XSUM so we discuss them separately in the following paragraphs.

% \kmnote{In order to line up with your later conclusion. I think you should say "have a moderate correlation with some aspects of human judgments". }
For CNN/DM, we observe that the reference-based automatic metrics have a moderate correlation with some aspects of human judgments, e.g., 
% \kmnote{I think you have to be clear here that this is for relevance, not across all aspects}
Rouge-L has a 0.72 Kendall's tau correlation coefficient with relevance in \Cref{tab:metrics}.
Such a level of correlation is comparable to that reported in \citet{fabbri2020summeval}, which measures the correlation of automatic metrics on evaluating finetuned LMs and even earlier neural summarization systems.
% \kmnote{Could you cut the rest of this on Fabbri? And for you conclusion, be specific that automatic metrics provide useful signals on relevance but not coherence. }
% For example, in the study of \citet{fabbri2020summeval}, the metric that correlates with relevance best is CHRF~\citep{Popovic2015chrFCN} and the Kendall's tau correlation is 0.58.
Therefore, we conclude that on CNN/DM automatic metrics can still provide useful signals in relevance.
% To contextualize these results, we compare to the automatic metrics evaluation on CNN/DM conducted by \citet{fabbri2020summeval}.
% On faithfulness, coherence, and relevance, the best metrics reported in \citet{fabbri2020summeval} achieve Kendall's tau correlation coefficients of 0.70, 0.39, and 0.58. \kmnote{This could use a bit more explanaation. I'm not sure if the comparison ends here or it really is in the next paragraph? Do you have a table of your results that you could point to from the first sentence? I would at least like to know your numbers. So I think you should be pointing to table 3? }

\begin{figure}
    \centering
    \includegraphics[width=\linewidth]{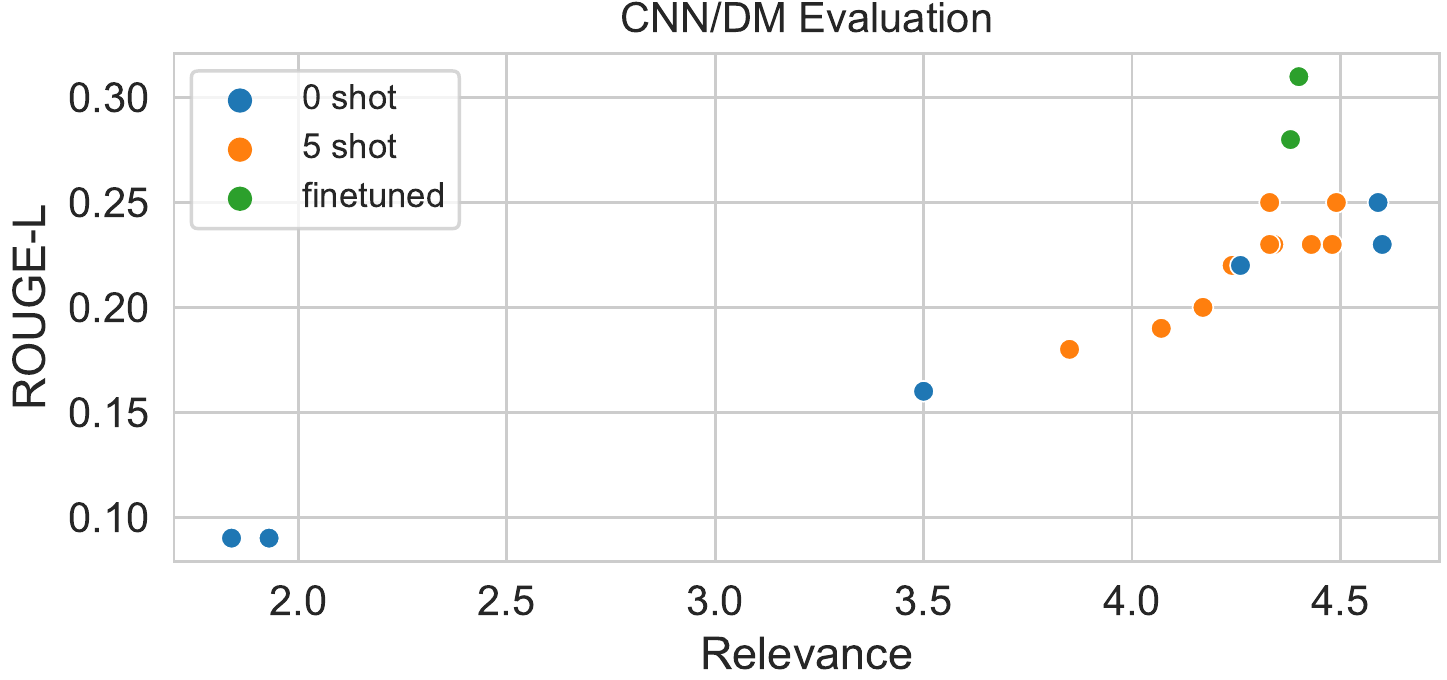}
    \caption{System-level Rouge-L vs. annotator rated relevance scores.}
    \label{fig:cnn_relevance_scatter}
\end{figure}

% \kmnote{This figure is missing so it's hard to see if it does help, but I find this paragraph particularly "waffling". Perhaps just skip this part and go right to the last paragraph where the opposite of correlation on CNN/DM holds. }
Studying the result more closely, we find that Rouge-L and human evaluation is more correlated when comparing within each model group.
We plot Rouge-L over the relevance rating in \Cref{fig:cnn_relevance_scatter} as an example.
First, we observe that Rouge-L does still prefer finetuned LMs (green points on top of the plots) to LLMs, consistent with prior work~\citep{gpt3-era}.
Despite this mistake, when only comparing LLMs with each other, we find that a larger than 0.05 Rouge-L difference usually translates to improved human evaluation.
% That said, Rouge-L is not effective in distinguishing among the best three to five systems in our study when the metric difference is small. 
% Rouge-L is able to distinguish better systems especially when the metric difference is larger than 0.05. \kmnote{This is a little unclear. Why do you say "however"? And why do you say "especially"? Do you mean "only"? }
% That said, Rouge-L is not effective in distinguishing among the best three to five systems in our study. 
% \kmnote{because why? Because the metric difference i s less than .05?}
% Besides Rouge-L, we see similar trends on other reference-based metrics. \kmnote{If you don't include them here, could you point to them in the appendix?}

On XSUM, the metrics have very low correlation with faithfulness and relevance but it is also because the reference summaries are terrible in these aspects~\citep[\Cref{tab:metrics}; also see][]{maynez-etal-2020-faithfulness}.
With such low-quality references, we do not expect reference-based metrics to extract useful information.
% but higher correlation with coherence. 
% We hypothesize that the different phenomenon may be attributed to the quality of reference summaries.
% \kmnote{Sentence below needs to be cleaned up. Perhaps "Reference-based metrics correlate better with human judgments on the aspects for which reference summaries also have better scores"}
% In contrast, reference-based metrics correlate well with coherence on XSUM, e.g. Rouge-L has a 0.71 kendal's tau cofficient.

Combining the results from two datasets, we find that reference-based metrics correlate better with human judgments on the aspects for which reference summaries also have better scores (e.g. CNN/DM relevance, XSUM coherence).
% \kmnote{should it be "the important role of good reference.."}
This points to the important role of quality reference summaries for reference-based metrics, as previously observed in machine translation~\citep{freitag-etal-2020-bleu}.

% The aspects that the reference-based metrics correlate better with human are the aspects that the reference summaries score better on in \Cref{tab:summarization-human-evaluation}: relevance for CNN/DM and coherence for XSUM.

% For XSUM we see that reference-based metrics have see a higher correlation with coherence that the other two dimensions. 
% On CNN/DM, however, coherence is the dimension along with metrics have the lowest correlation. Across both datasets we find that unlike human raters, automated metrics assign the highest scores to finetuned models. 
% When we look at the subset of systems without finetuned models (No FT), we see improved correlation numbers for the automated metrics. 
% \input{sections/metrics_analysis.tex}
\section{Comparing the Best LLM to Freelance Writers}
\label{sec:upwork_written_summaries}
In \Cref{sec:helm_benchmark}, we see that the low-quality reference summaries make studying and benchmarking LLMs difficult.
In this section, we address this by recruiting Upwork freelance writers to collect better quality summaries.
With this data, we aim to answer two important questions.
First, we would like to know whether the best LLM has reached human-level performance and how the summaries written by the best LLM differ from the ones written by humans. 
% shows that the best LLM has strong summarization performance.
Second, we want to understand how well reference-based metrics correlate with human judgments once we compute them with higher quality reference summaries.
% \tianyi{1. Models are better than Human 2. If so, why might they be. 3. Reference-based metrics can't be trusted. }
% In \Cref{sec:helm_benchmark}, we observe that the best LLM has become the state-of-the-art summarization model and performs much better than reference summaries.
% In this section, we aim to understand how does the best LLM compare to human writers and how the LLM generated summaries are different from the human written ones.
% \kmnote{Towards these goals implies you only got partway there. Could you say "To achieve these goals" or "To address these goals"? }
% Toward these goals, we conduct a pilot study to collect high-quality summaries through collaborating with freelance writers.
% \kmnote{Again, do you want to have your results up front? Slightly odd. And repetitive}
% We find that the best LLM is rated as good as the the freelance writers despite writing much more extractive summaries and 
% \kmnote{THis is not clear. How does the correlation improve better references???}
% that the correlation between metrics with human can improve with better references. 
% \kmnote{above make clear that it is the LLM that writes much more extractive summaries not the human}

\subsection{Experimental Setup}
In this section, we describe the process of recruiting summary writers and our summary writing instructions.

\paragraph{Data.}
For data used in our study, we select 50 articles from each of the CNN/DM and XSUM evaluation sets described in \Cref{sec:helm_experimental_setup} and assign each article to three writers.
For XSUM, we use the full articles rather than the preprocessed version where the first bolded sentence is removed.
% In terms of preprocessing, we note that in the XSUM articles, the first bolded sentences are treated as the reference summaries and removed, making the articles difficult to read. 
% Because we are not relying on the reference summaries in this study, we revert this process and place the first sentences back before we assign them to our summary writers.

\paragraph{Writer recruitment.}
% writer demographic
We recruit six writers who have had previous experience in writing blog posts, landing page introductions, or product descriptions from the freelance work platform Upwork.
% \kmnote{I find it difficult that you wait to refer to Upwork only here. Up above when you refer to "writer" I want ot know how you get writers. Of course, I do know, but the reader will not. }
After conducting a qualification round by asking writers to summarize five articles, we selected the best writers according to the faithfulness, coherence, and relevance of their summaries. 
Through an initial pilot study, we estimate that the time required to summarize a CNN/DM or XSUM article is around 12 to 15 minutes.
Therefore, we pay our writers \$4 for every article they summarize to following the recommended practice~\citep{Whiting2019FairWC}.
We based the assignments on writers' availability, with the most prolific writer summarizing 100 articles and the least prolific writer summarizing 35 articles.

\paragraph{Summary writing instructions.}
% instruction
For the annotation instruction, we instruct our writers to summarize each article in around 50 words\footnote{We conducted an initial study to pilot instructions and found that instructing writers with a sentence limit often resulted in summaries that differ significantly in length.}. 
% \kmnote{This is the first time you mention an initial study. You need to introduce or refer to it mroe as a parenthetical, perhaps a footnote. As a footnote you could say "We conducted an initial study to pilot instructions and found that ..". Or perhaps you could leave it that way in the main text.}
% Therefore, we instead provide a word limit and find that our writers are able to follow it well. 
To give better task grounding, we ask the writers to summarize as if they are writing a newletter to update their readers on news.  
We release the full annotation guideline along with our code release.
% Throughout the annotation process, we randomly inspect the summaries submitted by our writers and provide feedback. \kmnote{This is a little concerning. Did you use summaries generated before and after feedback? Or did you only use the ones they produced after feedback. What kind of feedback? Could this bias results ?}

% data
\paragraph{LLM Summaries Generation.}
Recently, \citet{Liu2022RevisitingTG} showed that length is a confounding factor in summarization human evaluation.
% length confound
To control this potential length confound, we modify the zero-shot prompt in \Cref{sec:helm_experimental_setup} to elicit summaries that are around 50 words, which is the same word limit provided to the freelance writers.
We found that the Instruct Davinci model consistently produces summaries that exceed a given word limit.
Therefore, we intentionally prompt the Instruct Davinci model with a 25 words limit to produce summaries with an average length of 50 words.
With this new prompt, we generate the summaries using the same hyperparameters described in \Cref{sec:helm_experimental_setup}.

\paragraph{Quality Control.}
\begin{table}[t]
    \centering
    \resizebox{\linewidth}{!}{
    \begin{tabular}{c|ccc}
    \toprule
     Model & Faithfulness & Coherence & Relevance \\
     \midrule
     Freelance Writer &  0.93 & 4.39 & 4.26 \\
     \midrule
     \Centerstack{Zero-shot\\Instruct Davinci} & 0.98 & 4.26 & 4.40 \\
     \midrule
     Reference Summaries & 0.64 & 3.59 & 3.45 \\
     \bottomrule
    \end{tabular}
    }
    \caption{Amazon Mechanical Turker evaluation results of the freelance writer summaries. Results of zero-shot Instruct Davinci and reference summaries are taken from \Cref{tab:summarization-human-evaluation} after averaging the corresponding ratings.}
    \label{tab:quality_control}
\end{table}
To verify the quality of the summaries written by freelance writers, we evaluate a random subset of 100 summaries using the same annotation scheme in \Cref{sec:helm_experimental_setup} using Mechanical Turkers. 
\Cref{tab:quality_control} reports the evaluation results, where we see that the freelance writer summaries have much higher quality than the original reference summaries in CNN/DM and XSUM.
In addition, we see that the difference between the freelance writer and Instruct Davinci in this evaluation is small. 
Next, we carry out more targeted evaluations to compare the summaries written by freelance writers and Instruct Davinci.

\subsection{Paired Comparison between LLM and Freelance Writers}
% Using this better set of human written summaries, we next benchmark the best LLM (Instruct Davinci) by having annotators directly compare the model generated summaries versus the human written summaries.

\label{sec:paired_eval_results}

\begin{figure}[t]
    \centering
    \includegraphics[width=\linewidth]{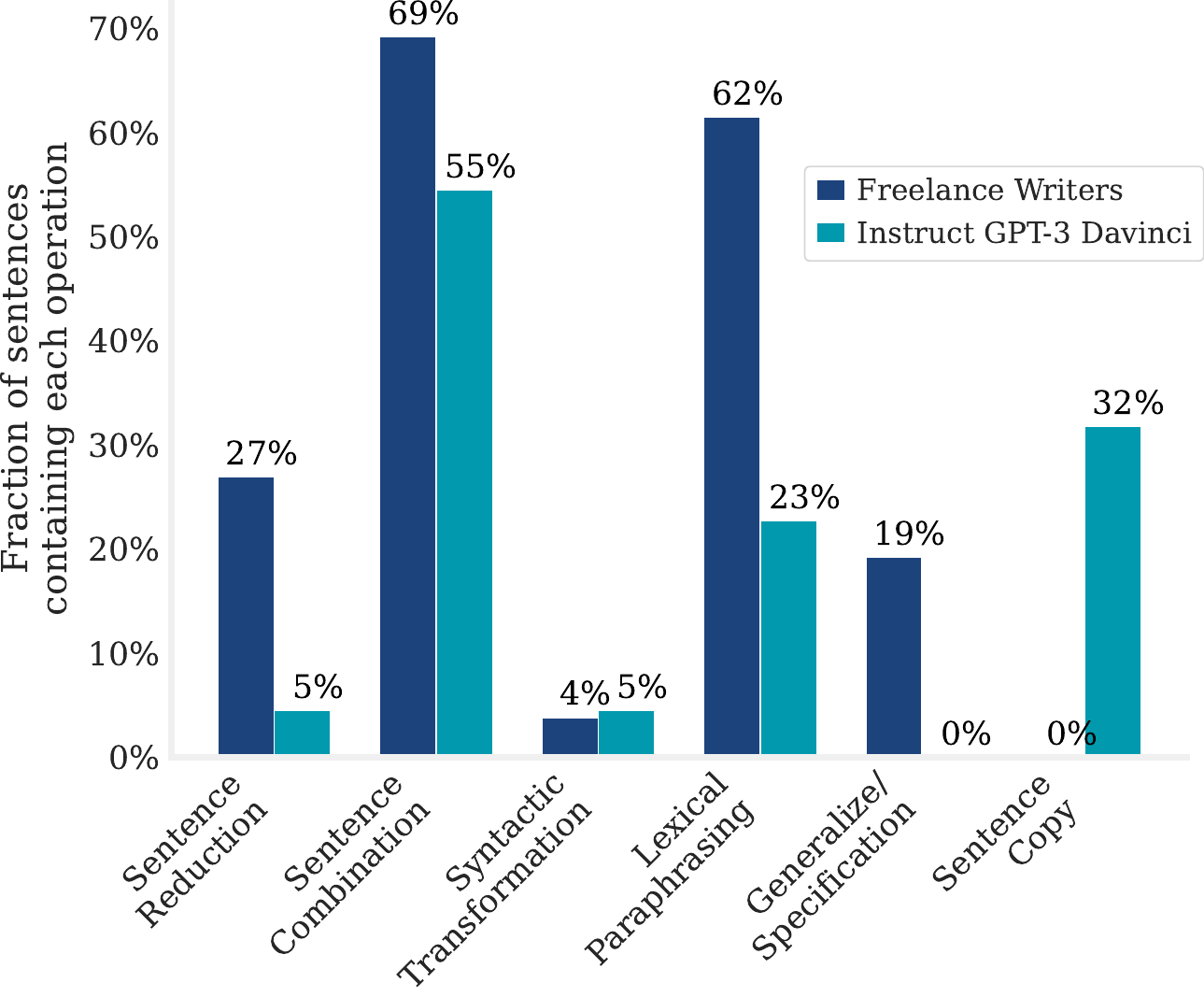}
    \caption{Distributions of cut and paste operations in the summaries written by freelance writers and by Instruct Davinci. By comparison, human written summaries contain more lexical paraphrasing and sentence reduction whereas the Instruct Davinci model has more direct copying from the article.}
    \label{fig:cut_and_paste_operations}
\end{figure}

\begin{figure*}
    \centering
    \includegraphics[width=\linewidth]{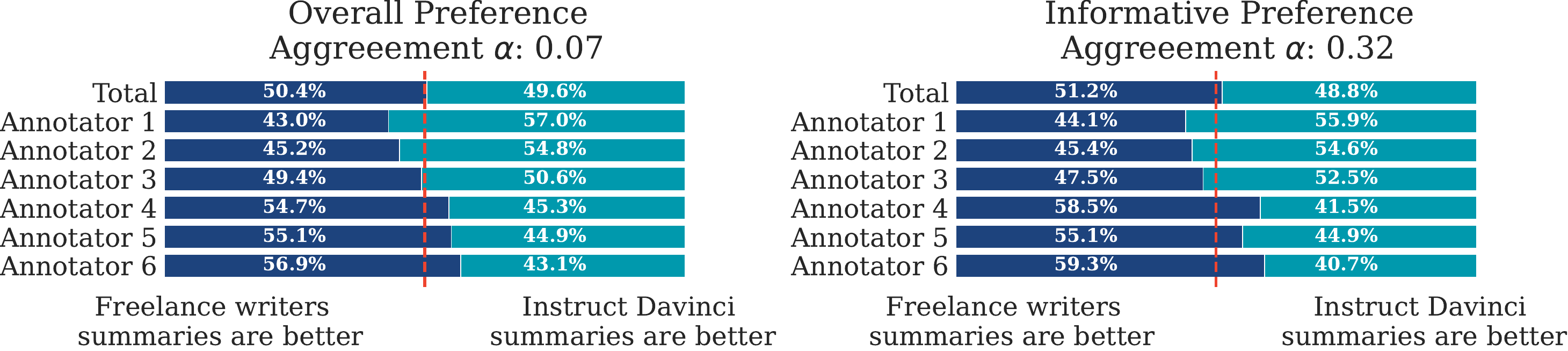}
    \caption{Human evaluation results comparing summaries written by freelance writers and summaries generated by Instruct GPT-3 Davinci. On aggregate, annotators equally prefer the freelance writers and Instruct Davinci. However, there is high variability in individual annotators' preferences. Notably, annotator 1 writes abstractive summaries but prefers the more extractive Instruct Davinci summaries.}
    \label{fig:human_preference}
\end{figure*}

\paragraph{Comparing Stylistic Differences.}
% \tianyi{justify why we look at extractiveness and abstractiveness by citing something.}
% We observe that summaries generated by Instruct Davinci are highly extractive whereas the summaries written by the freelance writers are much more abstractive.
Despite the similar performance in our quality control study, we find that LLM summaries and the freelance writer summaries have distinctive styles.
\Cref{fig:qualitative_example} shows an example summry written by the freelance writer.
Compared to the LLM generated summary, we find the freelance writer summary often contains more paraphrasing and copies less from the article.

To illustrate this stylistic difference, we measure two extractiveness measures, coverage and density, following \citet{Grusky2018NewsroomAD}.
Coverage is defined as the percentage of words in the summary that are also present in the article;
density is defined as the average length of the continuous text spans in the summary that are copied from the article.
Our analysis shows that the coverage and density for Instruct Davinci generated summaries are 0.92 and 12.1 whereas those for the writers written summaries are 0.81 and 2.07.
These measures show that the summaries generated by Instruct Davinci are highly extractive whereas the summaries written by the freelance writers are much more abstractive.

To have a finegrained understanding of these stylistic differences, we manually analyze the distrubution of ``cut and paste operations'' in these two sets of summaries.
\citet{Jing2000CutAP} identify a set of ``cut and paste'' operations for reusing text from the article, including sentence reduction, sentence combination, syntactic transformation, lexical paraphrasing, and generalization or specification. 
On top of these operations, we additionally include a sentence copy operation to account for summary sentences that are directly copied from the article.
Using this guideline, we manually annotate ten randomly sampled summary pairs written by Instruct Davinci and the freelance writers.

\Cref{fig:cut_and_paste_operations} reports the distribution of the cut and paste operations, showing the fraction of sentences that contain each operation.
First, we observe that the freelance writer summaries use lexical paraphrasing and generalization/specification much more frequently than the Instruct Davinci generated summaries.
Because both operations often involve using novel words that are not present in the article, this matches with the fact that the freelance writer summaries have lower coverage (0.81 vs. 0.92) than the Instruct Davinci summaries.
Second, we find that sentence combination is a common strategy used by both the freelance writers and Instruct Davinci.
% In particular, Instruct Davinci is able to string together segments copied from different article sentences coherently.
Third, we find that the freelance writers never copy 
an entire sentence directly from the article but Instruct Davinci does this more frequently.

In conclusion, we find that Instruct Davinci summarizes in a very different style than human writers.
We emphasize here that the freelance writers write in an abstractive style despite the fact that we have not explicitly instructed them to do so.
We also observe similarly abstractive styles across the six freelance writers.

\paragraph{Comparing Human Preference.}
We now return to our original goal of understanding wheter LLM generated summaries have quality
on par with the human written ones.
% \kmnote{"on-par" -> "on par" no hyphen}
% Given the distinctive styles of LLM summaries and freelance writer summaries, we want to understand which summaries are preferred by human. 
In the following paragraphs, we discuss our annotation design and recruitment process. 

% To understand the human preference, We instruct annotators to compare a pair of LLM summary and a freelance writer summary for the same article.
We conduct a blinded pairwise comparison evaluation between the best LLM Instruct Davinci and the freelance writers, similar to the evaluation in \citet{goyal-durrett-2020-evaluating}.
Besides selecting the better summary within each pair, the annotators can decide the summary pair to be equally good.
We release the full annotation instructions along with the code release for this project.
% Then, we ask the annotators to either select the better summary or decide that they are equally good.
% \kmnote{Give exact annotataion instructions in appendix and point to them.}
% In our instructions, we do not give away that one of the summary is written by the LLM and we randomize the order in which we show the summaries\footnote{We will release the annotation guideline after the TACL anonymous period.}. \kmnote{I am just seeing the footnote. When exactly would this be? And why do you wait? }

% \kmnote{awkward. How about "In order to compare the best LLM..."}
In order to compare the best LLM with the freelance writers, we annotate two aspects.
First, we solicit annotators' overall preference, which balances the multiple quality aspects such as faithfulness, coherence, and relevance.
Second, we solicit a more targeted measure of informativeness by asking the annotators to compare the number of facts in each summary.
For the informativeness measure, we are motivated by the hypothesis that a more abstractive writing style can pack more information into the summary given the same word count.
% Therefore, we added a second question where the annotator needs to identify which summary is more informative, defined as the summary containing the most amount of information.
While it is also interesting to compare summary coherence and relevance, we omit them because annotators were unable to differentiate these aspects from the overall preference in a pilot study.

% recruitment
For our recruitment process, we recruit five additional annotators through Upwork and retain one writer who participated in the previous round of summary writing\footnote{Other annotators left during the course of study due to change in freelance work schedule.}.
We carry out a qualification round and reject annotators whose ratings differ significantly from the authors' on a set of control questions for informativeness. 
% \kmnote{I don't understand this sentence. How did you get author ratings? Where do they come from? }
We give each annotator the same set of 100 summary pairs, where the average length of the freelance writer summaries and the Instruct Davinci summaries are 53.2 and 52.0 respectively.

\Cref{fig:human_preference} shows the results of the paired comparison. 
While we hypothesized that the more abstractive writing style can lead to more informative summaries, we do not find a significant effect in our annotator pool, 
who rate the more abstractive summaries to be more informative only 51.1\% of the time.
% \kmnote{You need to say which is higher. Presumably, abstractive is 51.2?}
% giving an informative preference of 51.2\% to 48.8\%. 
On the informative question, our annotators reached a moderate agreement (Krippendorff's alpha is 0.32), validating our annotation instruction and recruitment process.
Moving onto the more subjective overall preference, we find that our annotators equally prefer the freelance writer summaries and the Instruct Davinci summaries.
However, a closer analysis shows that there is significant variability in individual annotators' preference and the interannotator agreement is low (Krippendorff's alpha is 0.07).
This suggests that the quality of generated summaries is getting close to that of the freelance writer summaries and the comparison is dependent on each annotator's stylistic preference. 

One example of such stylistic preference is seen in the results from annotator 1, who also participated in the first round of summary writing.
Like other writers, annotator 1 summarizes in an abstractive style (2.5 density and 0.86 coverage).
However, annotator 1 prefers Instruct Davinci 57\% of the time even though it generated much more extractive summaries.
% Hence, we conclude that there could be a discrepancy between writing and rating summaries for annotators.
These results suggest an intriguing gap between annotator preferences when writing and evaluating summaries.

\subsection{Reevaluating Reference-based Metrics}
\label{sec:metric_2}
In \cref{sec:metric_1}, we saw that the performance of automated metrics may depend on the quality of reference summaries.
With the freelance writer summaries, we now conduct an initial study on the effect of using better quality summaries.
We focus on using Rouge-L for faithfulness evaluation on the XSUM dataset because the current reference summaries are known to be highly unfaithful~\citep{maynez-etal-2020-faithfulness}.

\begin{figure*}[htp]
    \centering
    \includegraphics[width=\textwidth]{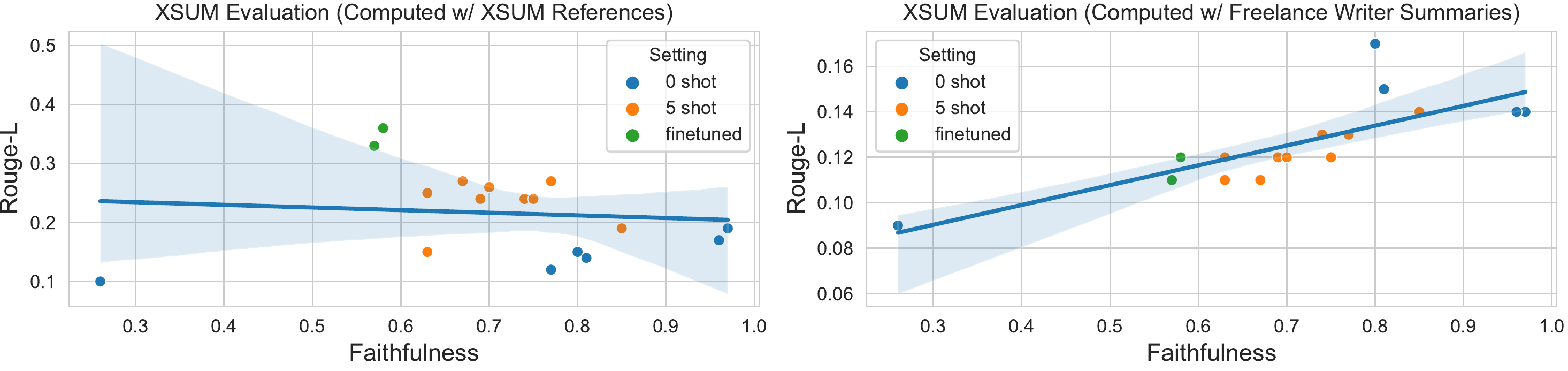}
    \caption{System-level Rouge-L vs. annotating rating of faithfulness. The left plot is computed with XSUM references, where the correlation is weak, and the right plot is computed with the freelance writer summaries, where the correlation is much improved.}
    \label{fig:xsum_faithfulness}
\end{figure*}

% don't work well for the XSUM dataset, in particular at evaluating faithfulness. One known issue with XSUM is that the references are poor proxies for summaries, and are usually not faithful to the input article \cite{maynez-etal-2020-faithfulness}. 

In \Cref{fig:xsum_faithfulness}, we plot the system-level Rouge-L against the human ratings.
The left plot shows results of computing Rouge-L with existing references summaries from XSUM, which has negative correlation with human ratings.
This result matches our expectation because the existing reference summaries are highly unfaithful.
On the right, we see the results of computing Rouge-L with the freelance writer summaries, which leads to a much more positive correlation.
% \kmnote{Another example of "waffling". I think you mean that there is only potential when correlation is high? But you could say it the opposite way in order to align with what you previously said: "We conclude that there is only potential to reference-based automatic evaluation when reference summary quality is high as shown by the high correlation here."}
Hence, we see that the usefulness of reference-based evaluation is closely linked to the quality of the references and we can improve metric correlation by using better reference summaries.
\section{Discussion}
% \kmnote{Should be "Implication for" not "Implication on".. throughout this section. }
\paragraph{Implication for model development.}
% Recent work such as \citet{gpt3-era} raised substantial interest in LLMs as an alternate, high-performance paradigm of automatic summarization.
% \kmnote{Unclear what you mean here. What is "strong performance against ..."?? do you mean "using high-quality annotators"? I think "poorly understood" may be seen as a strong criticism of Goyal. You refer to them a lot. Do you need to again here? Or could you say "the underlying reasons for their performance was not fully understood"}
% Although some LLMs display strong performance on par with freelance writers, the underlying reasons for their performance are not fully understood. 
% Some recent study such as \citet{gpt3-era} claim that ``GPT-3 represents a fundamental paradigm shift in summarization''. 
% While we share the excitement, we advocate for a more fine-grained understanding of the effectiveness of LLMs.
% We believe that there are many design decisions that go into the success of LLMs.
% In this study, we try to understand the worked toward such understanding by evaluating a diverse 
% \kmnote{"list" not the right word. Try "set" or "group"}
% set of LLMs and identified instruction tuning as the key contributing factor.
In this study, we build a systematic evaluation of a diverse set of LLMs and find that instruction tuning contributes the most to LLMs' summarization capability.
%KMFINAL2 ungrammatical
%We believe there are many research beyond 
We believe that there is much research beyond
our benchmarking effort that needs to be done to better understand the effect of instruction tuning.
Here we hypothesize three aspects that could account for the success of instruction tuning. 

First, the quality of the summariztion data used in instruction tuning can serve an important role.
Our findings in \Cref{sec:helm_benchmark} show that currently we are finetuning language models on low quality training data, which can account for their ineffectiveness.
At this point, we cannot rule out the possibility that when finetuned on higher quality data, finetuned LMs may perform much better.

Second, the learning algorithm used for instruction tuning can be important~\citep{instruct-gpt}.
While the exact training details are unknown, the success of Instruct Davinci might be credited to ``learning from human feedback''~\citep[LHF;][]{Stiennon2020LearningTS,Ziegler2019FineTuningLM}.
Contrary to supervised finetuning that trains systems on written summaries, learning from human feedback trains systems from binary labels of human preferences.
% \kmnote{Could you shorten this to "human preferences". }
% which generated summaries are preferred by annotators.
% RLHF trains a reward model to imitate human preference (e.g., binary label of which model generated summary is better) and uses the reward model to carry out reinforcement learning.
As we observe in \Cref{sec:paired_eval_results}, there is discrepancy in how annotators write and rate summaries.
% \kmnote{Make it more clear that this is an open quesiton. -> "While it is possible that LHF is superior to traditional supervised/finetuning precisely because of this discrepancy, more analysis is needed to validate this hypothesis." }
While it is possible that LHF have merits over the supervised learning/finetuning approach in exploiting this discrepancy, more analysis is needed to validate this hypothesis.
% \kmnote{But? Would be nice to have just a hint more of what you are thinking here}

Third, multi-task learning can be important. 
Instruct Davinci is trained on a diverse distribution of inputs and many previous studies have confirmed the effectiveness of multi-task learning. 
We look forward to understanding how summarization benefits from learning on other tasks.

\paragraph{Implication for Summarization Evaluation.}
Our work also reveals the difficulties in evaluating high-performance LLMs. 
As LLMs become increasingly close to human-level performance, human evaluation requires a larger number of samples and less noisy measurement to evaluate the quality of LLMs. 
Recently, \citet{Liu2022RevisitingTG} also point out the difficulties in conducting human evaluation for summarization and advocate using finegrained semantic units to match with reference summaries.
However, as our evaluation points out, not only are the existing reference summaries unreliable but the summaries written by well-paid freelance writers also may not outperform LLM summaries significantly.
Therefore, defining reference summaries as the ground truth may be overly restrictive as LLMs are approaching or even exceeding average human level performance.

% Our evaluation results in \Cref{sec:paired_eval_results} also indicate that there is inconsistency in how humans write and rate summaries. 
Not only is human evaluation limited by the reference quality, but it also is affected by the subjectivitiy in evaluation.
Individual variation shows that there are many acceptable ways to summarize and individuals may even show different preferences at different points in time (writing vs rating).
These factors in combination lead to the fact that we may have reached the limit of single document news summarization.
Existing benchmarks can still play a role in evaluating new models but only if evaluation is done correctly.
As LLMs improve, we believe that summarization can be better grounded in downstream applications where user values are better defined so that  annotators have a lower degree of freedom in balancing which quality aspects matter most to them.

\section{Conclusion}
% \kmnote{I think you could make the conclusion stronger. For example, you coudl say that high rouge score on XSUM is meaningless and can no longer be used as a claim that the model is better. You could say more about the contributing factors that  you did find and unresolved questions that remain open. If brave, you could say the single document news does provide a good testbed to evaluate evaluation practices, but given the individual preferences that you find, it may be that we have reached the limit with what we can do in this domain. }
In this work, we conducted a comprehensive human evaluation of ten LLMs, across the two most popular news summarization benchmarks. Through our experiments, we find that the state-of-the-art LLM performs on par with summaries written by freelance writers, with instruction tuning being the key factor for success.
%identified instruction tuning as the most important success factor, and also showed that the state-of-the-art LLM is comparable with Upwork freelance writers.
Beyond these findings, our work highlights the crucial role of good reference summaries in both summarization model development and evaluation. Unless the reference quality issue is addressed,
comparing zero-shot, few-shot, and finetuning performance will remain an open question, and the current benchmarks will provide limited value when used with reference-based evaluation.
Even when we address the quality issue and conduct a human evaluation with high-quality references, we observe a significant amount of individual variation from our annotator pool.
Due to these factors, evaluations for single document news summarization may be reaching their limits.
% Despite LLM's success, we believe there is ample room for future research to understand the contributing factors.
% Our work also highlighted the importance of quality reference summaries and we look forward to collecting quality training data to help compare different training paradigms.
\newpage
\section*{Acknowledgement}
This work is supported by an Open Philanthropy grant and partially supported by a gift from Northrup Grumman.
We thank the Stanford NLP group and the Stanford Center for Research on Foundation Models community for their feedback.

\bibliography{tacl2021}
\bibliographystyle{acl_natbib}
\end{document}